\documentclass[11pt,a4paper]{article}
\usepackage[hyperref]{acl2019}
\usepackage{times}
\usepackage{latexsym}
\usepackage{lipsum}
\usepackage{graphicx}
\usepackage{subcaption}
\usepackage{url}
\usepackage{float}
\usepackage{multicol}

\usepackage{tikz}

\aclfinalcopy 

\title{Visually grounded generation of entailments from premises}

\author{Somaye Jafaritazehjani$^{\dagger}$, Albert Gatt$^{\diamond}$, Marc Tanti$^{\diamond}$\\
  $^{\dagger}$IRISA, Universit\'{e} de Rennes I, France \\
  $^{\diamond}$Institute of Linguistics and Language Technology, University of Malta \\
  \texttt{somayeh.jafaritazehjani@irisa.fr}\\
  \texttt{\{albert.gatt, marc.tanti\}@um.edu.mt} \\
 }

\date{}

\begin{document}
\maketitle
\begin{abstract}
 Natural Language Inference (NLI) is the task of determining the semantic relationship between a premise and a hypothesis. In this paper, we focus on the {\em generation} of hypotheses from premises in a multimodal setting, to generate a sentence (hypothesis) given an image and/or
its description (premise) as the input. The main goals of this paper are (a) to investigate whether it is reasonable to frame NLI  as a generation task; and (b) to consider the degree to which grounding textual premises in visual information is beneficial to generation. We compare different neural architectures, showing through automatic and human evaluation that entailments can indeed be generated successfully. We also show that multimodal models outperform unimodal models in this task, albeit marginally. 
\end{abstract}

\section{Introduction}
\begin{figure*}
\centering
    \begin{subfigure}[b]{0.45\textwidth}
    \centering
        \includegraphics[scale=0.40]{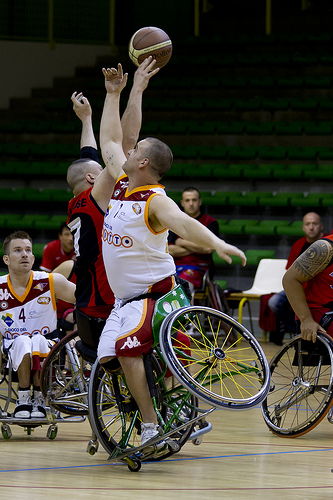}
        \caption{$P$: Four guys in wheelchairs on a basketball court two are trying to grab a ball in midair \\$Q$: Four guys are playing basketball.}\label{iamgesfor2premise1}
    \end{subfigure}
    \hfill
    \begin{subfigure}[b]{0.45\textwidth}
        \centering
        \includegraphics[scale=0.40]{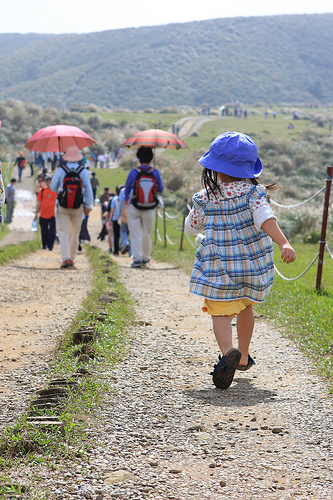}
        \caption{$P$: A little girl walking down a dirt road behind a group of other walkers\\$Q$: The girl is not at the head of the pack of people}
        \label{iamgesfor2premise2}
    \end{subfigure}  
    \caption{Images with premises and hypotheses} \label{iamgesfor2premise1and2}
\end{figure*}

Natural Language Inference (NLI) or Recognizing Textual Entailment (RTE) is typically formulated as a classification task: given a pair consisting of premise(s) P and hypothesis Q, the task is to determine if Q is entailed by P, contradicts it, or whether Q is neutral with respect to P \cite{dagan}. For example, in both the left and right panels of Figure \ref{iamgesfor2premise1and2}, the premise P in the caption entails the hypothesis Q.

In classical (i.e. logic-based) formulations \cite[e.g.,][]{coop:usin96}, P is taken to entail Q if Q follows from P in all `possible worlds'. Since the notion of `possible world' has proven hard to handle computationally, more recent approaches have converged on a probabilistic definition of the entailment relationship, relying on data-driven classification methods \cite[cf.][]{dagan,the-fourth-pascal-recognizing-textual-entailment-challenge}. In such approaches, P is taken to entail Q if, `typically, a nhuman reading P would infer that Q is most likely true' \cite{dagan}. Most approaches to RTE seek to identify the semantic relationship between P and Q based on their textual features. 

With a few exceptions \cite{Xie2019,Lai2018,Hoa}, NLI is defined in unimodal terms, with no reference to the non-linguistic `world'. This means that NLI models remain trapped in what \newcite{Roy2005} described as a `sensory deprivation tank', handling symbolic meaning representations which are ungrounded in the external world \cite[cf.][]{Harnad1990}. On the other hand, the recent surge of interest in NLP tasks at the vision-language interface suggests ways of incorporating non-linguistic information into NLI. For example, pairing the premises in Figure \ref{iamgesfor2premise1and2} with their corresponding images could yield a more informative representation from which the semantic relationship between premise and hypothesis could be determined. This would be especially useful if the two modalities contained different types of information, effectively making the textual premise and the corresponding image complementary, rather than redundant with respect to each other.

While NLI is a classic problem in Natural Language Understanding, it also has considerable relevance for generation. Many text generation tasks, including summarisation \cite{Nenkova2011}, paraphrasing \cite{androutsopoulos2010survey}, text simplification \cite{Siddharthan2014} and question generation \cite{piwek2012varieties}, depend on the analysis and understanding of an input text to generate an output that stands in a specific semantic relationship to the input. 

In this paper, we focus on the task of {\em generating} hypotheses from premises. Here, the challenge is to produce a text that follows from (`is entailed by') a given premise. Arguably, well-established tasks such as paraphrase generation and text simplification are specific instances of this more general problem. For example, in paraphrase generation, the input-output pair should be in a relationship of meaning preservation. 

In contrast to previous work on entailment generation \cite{Kolesnyk2016GeneratingNL,DBLP:journals/corr/StarcM16a}, we explore the additional benefits derived from grounding textual premises in image data. We therefore assume that the data consists of triples $\langle P, I, Q \rangle$, such that premises are accompanied by their corresponding images (I) as well as the hypotheses, as in Figure \ref{iamgesfor2premise1and2}. We compare multimodal models to a unimodal setup, with a view to determining to what extent image data helps in generating entailments.

The rest of this paper is structured as follows. In Section \ref{sec:motivation} we discuss further motivations for viewing NLI and entailment generation as multimodal tasks. Section \ref{sec:methodology} describes the dataset and architectures used; Section \ref{sec:results} presents experimental results, including a human evaluation. Our conclusion (Section \ref{sec:conclusion}) is that it is feasible to frame NLI as a generation task and that incorporating non-linguistic information is potentially profitable. However, in line with recent research evaluating Vision-Language (VL) models \cite{Shekhar2017,Wang2018,Hoa,sivl2019sensitivity}, we also find that current architectures are unable to ground textual representations in image data sufficiently.

\section{NLI, Multimodal NLI and Generation}\label{sec:motivation}
Since the work of \newcite{dagan}, research on NLI has largely been data-driven \cite[see][for an overview]{Sammons2012}. Most recent approaches rely on neural architectures \cite[e.g.][]{Rocktschel2015ReasoningAE,Chen2016,Wang2017}, a trend which is clearly evident in the submissions to recent RepEval challenges \cite{Nangia2017}. 

Data-driven NLI has received a boost from the availability of large datasets such as SICK \cite{sick}, SNLI \cite{SNLI} and MultiNLI \cite{Williams2018}. Some recent work has also investigated multimodal NLI, whereby the classification of the entailment relationship is done on the basis of image features \cite{Xie2019,Lai2018}, or a combination of image and textual features \cite{Hoa}. In particular, \newcite{Hoa} exploited the fact that the main portion of SNLI was created by reusing image captions from the Flickr30k dataset \cite{youn:from14}  as premises, for which entailments, contradictions and neutral hypotheses were subsequently crowdsourced via Amazon Mechanical Turk \cite{SNLI}. This makes it possible to pair premises with the images for which they were originally written as descriptive captions, thereby reformulating the NLI problem as a Vision-Language task. There are at least two important motivations for this:

\begin{enumerate}
\item The inclusion of image data is one way to bring together the classical and the probabilistic approaches to NLI, whereby the image can be viewed as a (partial) representation of the `world' described by the premise, with the entailment relationship being determined jointly from both. This is in line with the suggestion by \newcite{youn:from14}, that images be considered as akin to the `possible worlds' in which sentences (in this case, captions) receive their denotation.

\item A multimodal definition of NLI also serves as a challenging testbed for VL NLP models, in which there has been increasing interest in recent years. This is especially the case since neural approaches to fundamental computer vision (CV) tasks have yielded significant improvements \cite{LeCun2015}, while also  making it possible to use pre-trained CV models in multimodal neural architectures, for example in tasks such as image captioning \cite{Bernardi:2016:ADG:3013558.3013571}. However, recent work has cast doubt on the extent to which such models are truly exploiting image features in a multimodal space \cite{Shekhar2017,Wang2018,sivl2019sensitivity}. Indeed, \newcite{Hoa} also find that image data contributes less than expected to determining the semantic relationship between premise-hypothesis pairs in the classic RTE labelling task. 
\end{enumerate}

There have also been a few approaches to entailment generation, which again rely on the SNLI dataset. \newcite{Kolesnyk2016GeneratingNL} employed a sequence-to-sequence architecture with attention \cite{Sutskever2014,Bahdanau2015} to generate hypotheses from premises. They extended this framework to the generation of inference chains by recursively feeding the model with the generated hypotheses as the input. \newcite{DBLP:journals/corr/StarcM16a} proposed different generative neural network models to generate a stream of hypotheses given $\langle$premise, Label$\rangle$ pairs as the input.

In seeking to reframe NLI as a generation task, the present paper takes inspiration from these approaches. However, we are also interested in framing the task as a multimodal, VL problem, in line with the two motivations noted at the beginning of this section. Given that recent work has suggested shortcomings in the way VL models adequately utilise visual information, a focus on multimodal entailment generation is especially timely, since it permits direct comparison between models utilising unimodal  and multimodal input.

As Figure \ref{iamgesfor2premise1and2} suggests, given the triple $\langle P, I, Q \rangle$, the hypothesis Q could be generated from the premise $P$ only, from the image $I$, or from a combination of $P+I$. Our question is whether we can generate better entailments from a combination of both, compared to only one of these input modalities. 



\section{Methodology}\label{sec:methodology}

\begin{figure*}[t]
    \centering
    \includegraphics[width=130mm]{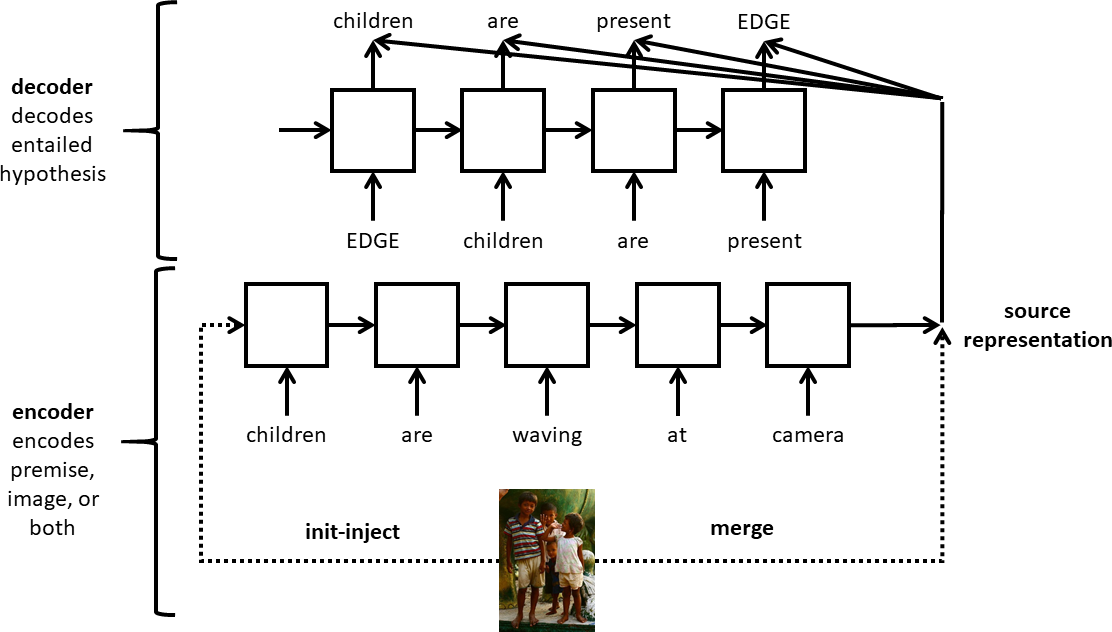}
    \caption{Architecture schema that is instantiated in the different models. The encoder returns a source representation which conditions the decoder into generating the entailed hypothesis. The encoder of multimodal models either (1) initialises the encoder RNN with the image features (init-inject) and returns the final state; or (2) concatenates the encoder RNN's final state with the image features (merge) and returns the result; or (3) returns the image features as is without involving the encoder RNN. The unimodal model encoder returns the encoder RNN's final state as is. The source representation is concatenated to the decoder RNN's states prior to passing them to the softmax layer.}
    \label{multimodal-se-seg_general}
\end{figure*}

\subsection{Data}\label{sec:data}
We focus on the subset of entailment pairs in the SNLI dataset \cite{SNLI}. The majority of instances in SNLI consist of premises that were originally elicited as descriptive captions for images in Flickr30k~\cite{youn:from14,flickr}.\footnote{A subset of 4000 cases in SNLI was extracted from the VisualGenome dataset \cite{Krishna2017}. Like \newcite{Hoa}, we exclude these from our experiments.}

In constructing the SNLI dataset, Amazon Mechanical Turk workers were shown the captions/premises without the corresponding images, and were asked to write a new caption that was (i) true, given the premise (entailment); (ii) false, given the premise (contradiction); and (iii) possibly true (neutral).

Following \newcite{Hoa}, we use a multimodal version of SNLI, which we refer to as V-SNLI, which was obtained by mapping SNLI premises to their corresponding Flickr30k images \cite[][use a similar approach]{Xie2019}. The resulting V-SNLI therefore contains premises which are assumed to be true of their corresponding image. Since we focus exclusively on the entailment subset, we also assume, given the instructions given to annotators, that the hypotheses are also true of the image. Figure \ref{iamgesfor2premise1and2} shows two examples of images with both premises and hypotheses.

Each premise in the SNLI dataset includes several reference hypotheses by different annotators.
In the original SNLI train/dev/test split, some premises show up in both train and test sets, albeit with different paired hypotheses.
For the present work, once premises were mapped to their corresponding Flickr30k images, all premises corresponding to an image were grouped, 
and the dataset was resplit so that the train/test partitions had no overlap, 
resulting in  a split of 182,167 (train), 3,291 (test) and 3,329 (dev) entailment pairs. This also made it possible to use a multi-reference approach to evaluation (see Section~\ref{sec:eval}).

Prior to training, all texts were lowercased and tokenized.\footnote{Experiments without lowercasing showed only marginal differences in the results.} Vocabulary items with frequency less than 10 were mapped to the token $\langle UNK \rangle$. The unimodal model (see Section \ref{sec:models}) was trained on textual P-Q pairs from the original SNLI data, while the multimodal models were trained on pairs from V-SNLI where the input consisted of P+I (for the multimodal text+image model), or I only (for the image-to-text model), to generate entailments.

\subsection{Models}\label{sec:models}
Figure~\ref{multimodal-se-seg_general} is an illustration of the architecture schema that is employed for both multimodal and unimodal architectures. A sequence-to-sequence encoder-decoder architecture is used to predict the entailed hypothesis from the source. The source can be either a premise sentence, an image, or both. 

Both the encoder and the decoder use a 256D GRU RNN and both embed their input words using a 256D embedding layer. In multimodal models, the encoder either produces a multimodal source representation that represents the sentence-image mixture or produces just an image representation. In the unimodal model, the encoder produces a representation of the premise sentence only. 

This encoded representation (multimodal or unimodal) is then concatenated to every state in the decoder RNN prior to sending it to the softmax layer, which will predict the next word in the entailed hypothesis.

All multimodal models extract image features from the penultimate layer of the pre-trained VGG-19 convolutional neural network \citep{Simonyan2014}. All models were implemented in Tensorflow, and trained using the the Adam optimizer~\citep{adam} with a learning rate of .0001 and a cross-entropy loss.

The models compared are as follows:

\begin{enumerate}
\item {\bf Unimodal} model: A standard sequence-to-sequence architecture developed for neural machine translation (NMT).  It uses separate RNNs for encoding the source text and for decoding the target text. This model encodes the premise sentence alone and generates the entailed hypothesis. In the unimodal architecture, the final state of the encoder RNN is used as the representation of the premise P.

\item Multimodal, text+image input, init-inject ({\bf T+I-Init}): A multimodal model in which the image features are incorporated at the encoding stage, that is, image features are used to initialise the RNN encoder. This architecture, which is widely used in image captioning \citep{Devlin2015,Liu2016} is referred to as  {\em init-inject}, following \citet{DBLP:tanti/corr/abs-1708-02043-0}, on the grounds that image features are directly injected into the RNN.

\item Multimodal, text+image input, merge ({\bf T+I-Merge}): A second multimodal model in which the image features are concatenated with the final state of the encoder RNN. Following \citet{DBLP:tanti/corr/abs-1708-02043-0}, we refer to this as the {\em merge} model. This too is adapted from a widely-used architecture in image captioning \citep{Mao2014,Mao2015,Mao2015a,Hendricks2016}. The main difference from {\em init-inject} is that here, image features are combined with textual features immediately prior to decoding at the softmax layer, so that the RNN does not encode image features directly. 

\item Multimodal, image input only ({\bf IC}): This model is based on a standard image captioning setup \citep{Bernardi:2016:ADG:3013558.3013571} in which the entailed hypothesis is based on the image features only, with no premise sentence. This architecture leaves out the encoder RNN completely. Note that, while this is a standard image captioning setup, it is put to a somewhat different use here, since we do not train the model to generate captions (which correspond to the premises in V-SNLI) but the hypotheses (which are taken to follow from the premises).
\end{enumerate}

\begin{table*}[!h]
\centering
\begin{tabular}{|l |l |l| l|l|}
\hline
\textbf{Model}& \textbf{BLEU-1} & \textbf{METEOR} & \textbf{CIDEr} & \textbf{Perplexity} \\
\hline
Unimodal& \textbf{0.695} & 0.267&0.938 & \textbf{7.23}\\
\hline
T+I-Init &0.634 &0.239 &0.763&10.73\\
\hline
T+I-Merge &0.686 & \textbf{0.271}&\textbf{0.955} &7.26 \\
\hline
IC &0.474 & 0.16& 0.235 & 13.7  \\
\hline
\end{tabular}
\caption{Automatic evaluation results for unimodal and multimodal models}
\label{results1}
\end{table*}

\subsection{Evaluation metrics}\label{sec:eval}
We employed the evaluation metrics BLEU-1~\citep{Papineni2002}, METEOR~\citep{Lavie2007} and CIDEr~\citep{Vedantam2015} to compare generated entailments against the gold outputs in the model. We also compare the model perplexity on the test set. 

As noted earlier, we adopted a multi-reference approach to evaluation, exploiting the fact that in our dataset we  grouped all reference hypotheses corresponding to each premise (and its corresponding image). Thus, evaluation metrics are calculated by comparing the generated hypotheses to a group of reference hypotheses. This is advantageous, since n-gram based metrics such as BLEU and METEOR are known to yield more reliable results when multiple reference texts are available.

\section{Results}\label{sec:results}

\subsection{Metric-based evaluation}\label{sec:metrics}
The results obtained by all models are shown in Table~\ref{results1}. The T+I-Merge architecture outperforms all other models on CIDEr and METEOR, while the unimodal model, relying only on textual premises, is marginally better on BLEU and has slightly lower perplexity. 

The lower perplexity of the unimodal model is is unsurprising given that in the unimodal model, the decoder is only conditioned on textual features without images. 
However, it should also be noted that the unimodal model also fares quite well on the other two metrics, and is only marginally worse than T+I-Merge. Its score on BLEU-1 is also competitive with that reported by \newcite{Kolesnyk2016GeneratingNL} for their seq2seq model, which obtained a BLEU score of 0.428. Note, however, that this score was not obtained using a multi-reference approach, that is, each generated candidate was compared to each reference candidate, rather than the set of references together. Recomputing our BLEU-1 score for the Unimodal model in this way, we obtain a score of 0.395, which is marginally lower than that reported by \citet{Kolesnyk2016GeneratingNL}.

The IC model, which generates entailments exclusively from images, ranks lowest on all metrics. This is probably due to the fact that, contrary to the usual setup for image caption generation, this model was trained on image-entailment pairs rather than directly on image-caption pairs. Although, based on the instructions given to annotators \cite{SNLI}, entailments in SNLI are presumed to be true of the image, it must be noted that the entailments were not elicited as descriptive captions with reference to an image, unlike the premises.

We discuss these results in more detail in Section \ref{sec:analysis}, after reporting on a human evaluation.

\subsection{Human evaluation}

We conducted a human evaluation experiment, designed to address the question whether better entailments are generated from textual premises, images, or a combination of the two. In the metric-based evaluation (Section \ref{sec:metrics}), the best multimodal model was T+I-Merge; hence, we use the outputs from this model, comparing them to the outputs from the unimodal and the image-only IC model.

\paragraph{Participants} Twenty self-reported native or fluent speakers of English were recruited through social media and the authors' personal network.

\paragraph{Materials} A random sample of 90 instances from the V-SNLI test set was selected. Each consisted of a premise, together with an image and three entailments generated using the Unimodal, IC and I+T-Merge models, respectively. The 90 instances were randomly divided into three groups. Participants in the evaluation were similarly allocated to one of three groups. The groups were rotated through a latin square so that each participant saw each of the 90 instances, one third in each of the three conditions (Unimodal, IC or I+T-Merge). Equal numbers of judgments were thus obtained for each instance in each condition, while participants never saw a premise more than once. 

\begin{figure}
    \centering
    \includegraphics[scale=0.4]{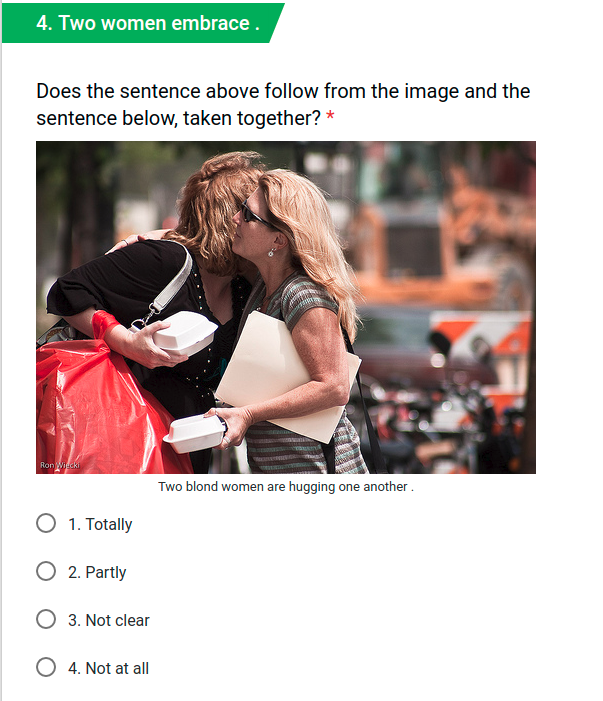}
    \caption{Example of an evaluation test item, with a sentence generated from both text and image.}
    \label{fig:eval}
\end{figure}

\paragraph{Procedure} Participants conducted the evaluation online. Test items were administered in blocks, by condition (text-only, image-only or text+image) and presented in a fixed, randomised order to all participants. However, participants evaluated items in the three conditions in different orders, due to the latin square rotation. Each test case was presented with an input consisting of textual premise, an image, or both. Participants were asked to judge to what extent the generated sentence followed from the input. Answers were given on an ordinal scale with the following qualitative responses: {\em totally}; {\em partly}; {\em not clear} or {\em not at all}. Figure \ref{fig:eval} shows an example with a generated entailment in the text+image condition.

\begin{figure}
    \centering
    \includegraphics[scale=0.55]{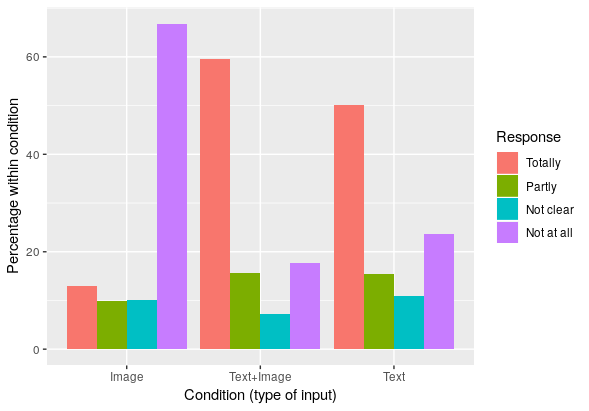}
    \caption{Proportion of responses in each category for each input condition.}
    \label{fig:eval-results}
\end{figure}

\paragraph{Results} Figure \ref{fig:eval-results} shows the proportion of responses in each category for each of the three conditions. The figure suggests that human judgments were in line with the trends observed in Section \ref{sec:metrics}. More participants judged the generated entailments as following totally or partly from the input, when the input consisted of text and image together (T+I-Merge), followed closely by the case where the input was text only (Unimodal). By contrast, in the case of the image-only (IC) condition, the majority of cases were judged as not following at all. 

Possible reasons for these trends are discussed in the following sub-section in light of further analysis. First, we focus on whether the advantage of incorporating image features with text is statistically reliable. We coded responses as a binomial variable, distinguishing those cases where participants responded with `totally' (i.e. the generated entailment definitely follows from the input) from all others. Note that this makes the evaluation conservative, since we focus only on the odds of receiving the most positive judgment in a given condition. 

We fitted logit mixed-effects models. All models included random intercepts for participants and items (premise IDs) and random slopes for participants and items by condition. Where a model did not converge, we dropped the by-items random slope term. All models were fitted and tested using the {\tt lme4} package in R \cite{Bates2014}. 

The difference between the three types of input Condition significantly affected the odds of a positive (`totally') response ($z=8.06, p < .001$). 

We further investigated the impact of incorporating image features with or without text through planned comparisons. A mixed-effects model comparing the image-only to the text+image condition showed that the latter resulted in significantly better output as judged by our participants ($z=7.80, p < .001$). However, the text+image model did not significantly outperform the text-only unimodal model, despite the higher percentage of positive responses in Figure \ref{fig:eval-results} ($z=1.61, p > .1$). This is consistent with the CIDEr and METEOR scores, which show that T+I-Merge outperforms the Unimodal model, but only marginally.

\subsection{Analysis and discussion}\label{sec:analysis}

\begin{table*}[t]
  \centering
   \begin{tabular}{|l|l| }
    \hline
\textbf{Source}  &\textbf{Sentence}    \\ \hline
P (Figure~\ref{iamgesfor2premise1}) & Four guys in wheelchairs on a basketball court two are trying to grab a ball in midair.\\
Q (ref)  & Four guys are playing basketball. \\
\hline
IC & A man is playing a game. \\
T+I-Init & Two men are jumping. \\
T+I-Merge &  The basketball players are in the court.\\
Unimodal & The men are playing basketball.    \\ 
\hline      
\hline
P (Figure~\ref{iamgesfor2premise2}) & A little girl walking down a dirt road behind a group of other walkers.\\
Q (ref) & The girl is not at the head of the pack of people. \\
\hline
IC & A man is wearing a hat. \\
T+I-Init & A young girl is walking down a path. \\
T+I-Merge &  A group of people are walking down a road.\\
Unimodal &  A girl is walking .    \\ 
\hline      
  \end{tabular}
  \caption{Premises and reference captions, with output examples. These test cases correspond to the images in Figure~\ref{iamgesfor2premise1and2}.}
  \label{allmodelsoutputs}
\end{table*}

Two sets of examples of the outputs of different models are provided in Table~\ref{allmodelsoutputs}, corresponding to the images with premises and entailments shown in Figure~\ref{iamgesfor2premise1}. 

These examples, which are quite typical, suggest that the IC model is simply generating a descriptive caption which only captures some entities in the image. We further noted that in a number of cases, the IC model also generates repetitive sentences that are not obviously related to the image. These are presumably due to predictions relying extensively on the language model itself, yielding stereotypical `captions' which reflect frequent patterns in the data. Together, these reasons may account for the poor performance of the IC model on both the metric-based and the human evaluations.

The main difference between the Unimodal and the multimodal T+I models seems to be that they generate texts that correspond to different, but arguably valid, entailments of the premise. However, the fact remains that the ungrounded Unimodal model yields good results that are close to those of T+I-Merge.
While humans judged the T+I-Merge entailments as better than the unimodal ones overall, this is not a significant difference, at least as far as the highest response category (`totally') is concerned.

These limitations of multimodal models echo previous findings. In image captioning, for example, it has been shown that multimodal models are surprisingly insensitive to changes in the visual input. For instance \citet{Shekhar2017} found that VL models perform poorly on a task where they are required to distinguish between a `correct' image (e.g. one that corresponds to a caption) and a completely unrelated foil image \cite{Shekhar2017}. Similarly, \newcite{sivl2019sensitivity} find that image captioning models exhibit decreasing sensitivity to the input visual features as more of the caption is generated. In their experiments on multimodal NLI, \newcite{Hoa} also found that image features contribute relatively little to the correct classification of pairs as entailment, contradiction or neutral. These results suggest that current VL architectures do not exploit multimodal information effectively.

Beyond architectural considerations, however, there are also properties of the dataset which may account for why linguistic features play such an important role in generating hypotheses. We discuss two of these in particular.

\paragraph{V-SNLI is not (quite) multimodal} SNLI hypotheses were elicited from annotators in a unimodal setting, without reference to the images. Although we believe that our assumption concerning the truth of entailments in relation to images holds (see Section \ref{sec:data}), the manner in which the data was collected potentially results in hypotheses which are in large measure predictable from linguistic features, accounting for the good performance of the unimodal model. If correct, this also offers an explanation for the superior performance of the T+I-Merge model, compared to T+I-Init. In the former, we train the encoder language model separately, mixing image features at a late stage, thereby allocating more memory to linguistic features in the RNN, compared to T+I-Init, where image features are used to initialise the encoder. Apart from yielding lower perplexity (see Table~\ref{results1}), this allows the T+I-Merge model to exploit linguistic features to a greater extent than is possible in T+I-Init. 

\paragraph{Linguistic biases} Recently, a number of authors have expressed concerns that NLI models may be learning heuristics based on superficial syntactic features rather than classifying entailment relationships based on a deeper `understanding' of the semantics of the input texts \cite{McCoy2019}.
Indeed, studies on the SNLI dataset have shown that it contains several linguistic biases \cite{Gururangan2018}, such that the semantic relationship (entailment/contradiction/neutral) becomes predictable from textual features of the hypotheses alone, without reference to the premise. \newcite{Gururangan2018} identify a `hard' subset of SNLI where such biases are not present.

Many of these biases are due to a high degree of similarity between a premise and a hypothesis. For example, contradictions were often formulated by annotators by simply including a negation, while entailments are sometimes substrings of the premises, as in the following pair:

\begin{itemize}
    \item[P]: A bicyclist riding down the road wearing a helmet and a black jacket.
    \item[Q]:  A bicyclist riding down the road.
\end{itemize}

Such biases could also account for the non-significant difference in performance between the two best models, Unimodal and T+I-Merge (which also allocates all RNN memory to textual features, unlike T+I-Init). In either case, it is possible that a hypothesis is largely predictable from the textual premise, irrespective of the image, and generation boils down to `rewriting' part of the premise to produce a similar string (see, e.g., the last example in Table \ref{allmodelsoutputs}). 

\begin{figure*}
\centering
    \begin{subfigure}[b]{0.45\textwidth}
    \centering
        \includegraphics[scale=0.5]{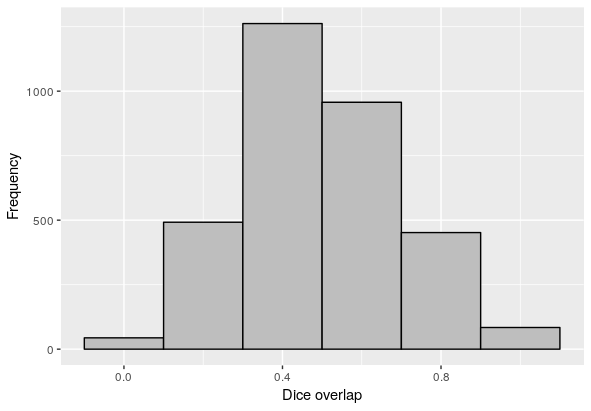}
        \caption{Histogram of Dice overlap values}\label{fig:dicehist}
    \end{subfigure}
    \hfill
    \begin{subfigure}[b]{0.45\textwidth}
        \centering
        \includegraphics[scale=0.5]{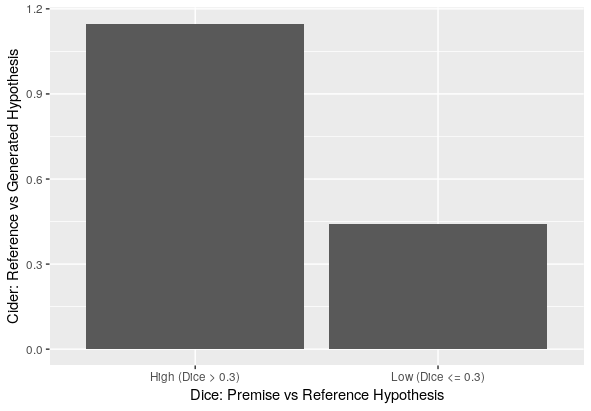}
        \caption{CIDEr scores for Low vs High overlap test cases}
        \label{fig:dicecider}
    \end{subfigure}  
    \caption{Dice overlap vs CIDEr scores (T+I-Merge)} \label{fig:overlap}
\end{figure*}

If textual similarity is indeed playing a role, then we would expect a model to generate entailments ($Q_{gen}$) with a better CIDEr score, in those cases where there is a high degree of overlap between the premise and the reference hypothesis ($Q_{ref}$). 

We operationalised overlap in terms of the Dice coefficient\footnote{$$Dice(P,Q_{ref}) = \frac{2 \times \vert P \cap Q_{ref} \vert}{\vert P \vert + \vert Q_{ref} \vert}$$}, computed over the sets of words in $P$ or $Q_{ref}$, after stop word removal.\footnote{Stopwords were removed using the built-in English stopword list in the Python NLTK library.} As shown by the histogram in Figure~\ref{fig:dicehist}, a significant proportion of the $P$-$Q_{ref}$ pairs have a relatively high Dice coefficient ranging from 0.4 to 0.8. We divided test set instances into those with `low' overlap (Dice $\leq$ 0.3; $n = 536$) and those with high overlap (Dice $>$ 0.3; $n=2755$). Figure \ref{fig:dicecider} displays the mean CIDEr score between entailments generated by the T+I-Merge model, and reference entailments, as a function of whether the reference entailment had high overlap with the hypothesis.

Clearly, for those cases where the overlap between $P$ and $Q_{ref}$ was high, the T+I-Merge model obtains a higher CIDEr score between generated and reference outputs. This is confirmed by Pearson correlation coefficients between Dice coefficient and CIDEr, computed for each of the high/low overlap subsets: On the subset with high Dice overlap, we obtain a significant positive correlation ($r=0.20, p < .001$); on the subset with low overlap, the correlation is far lower and does not reach significance ($r=.06, p > .1$).

\section{Conclusion}\label{sec:conclusion}
This paper framed the NLI task as a generation task and compared the role of visual and linguistic features in generating entailments. 
To our knowledge, this was the first systematic attempt to compare unimodal and multimodal models for entailment generation. 

We find that grounding entailment generation in images is beneficial, but linguistic features also play a crucial role. Two reasons for this may be adduced. On the one hand, the data used was not elicited in a multimodal setting, despite the availability of images. It also contains linguistic biases, including relatively high degrees of similarity between premise and hypothesis pairs, which may result in additional image information being less important. On the other hand, it has become increasingly clear that Vision-Language NLP models are not grounding language in vision to the fullest extent possible. The present paper adds to this growing body of evidence.

Several avenues for future work are open, of which three are particularly important. First, in order to properly assess the contribution of grounded language models for entailment generation, it is necessary to design datasets in which the textual and visual modalities are complementary, rather than redundant with respect to each other. Second, while the present paper focused exclusively on entailments, future work should also consider generating contradictions. Finally, further research should investigate more sophisticated Vision-Language architectures, especially incorporating attention mechanisms.

\section*{Acknowledgments}
The first author worked on this project while at the University of Malta, on the Erasmus Mundus European LCT Master Program. The third author was supported by the Endeavour Scholarship Scheme (Malta). We thank Raffaella Bernardi, Claudio Greco and Hoa Vutrong and our anonymous reviewers for their helpful comments.

\bibliography{inlg2019multimodalgen}
\bibliographystyle{acl_natbib}
\end{document}